\documentclass[letterpaper, 10 pt, conference]{ieeeconf}  

\IEEEoverridecommandlockouts                              

\usepackage{amsmath} 
\usepackage{amssymb}  

\usepackage{lipsum} 
\usepackage{xcolor,graphicx}
\usepackage{caption}
\usepackage[position=top]{subfig}

\newcommand{\p}{\partial}

\graphicspath{{./}{./figs/}}

\newif\ifcomments
\commentstrue

\title{\LARGE \bf
Coarse-grained and Emergent Distributed Parameter Systems \\from Data*
}

\author{Hassan Arbabi$^{1}$, Felix P. Kemeth$^{2}$, Tom Bertalan$^{2}$ and Ioannis Kevrekidis$^{2}$
\thanks{*This work was partially supported by DARPA and by the A.R.O. through a MURI program.}
\thanks{$^{1}$H.A. is with the Department of Mechanical Engineering, MIT, Cambridge, MA 02139, USA.}%
\thanks{$^{2}$F.K., T.B and I.K. (corresponding author, {\tt\small yannisk@jhu.edu}) are with the Department of Chemical and Biomolecular Engineering at Johns Hopkins University
        Baltimore, MD 21218, USA.  }%
}

\begin{document}

\maketitle
\thispagestyle{empty}
\pagestyle{empty}

\begin{abstract}

We explore the derivation of distributed parameter system evolution laws 
(and in particular, partial differential operators and associated partial differential equations, PDEs) 
from spatiotemporal data.
This is, of course, a classical identification problem; our focus here is on the use of manifold
learning techniques (and, in particular, variations of Diffusion Maps) in conjunction 
with neural network learning algorithms that allow us to attempt this task when 
the dependent variables, and even the independent variables of the PDE are not known \emph{a priori}
and must be themselves derived from the data.
The similarity measure used in Diffusion Maps for {\em dependent} coarse variable detection involves distances between local particle distribution observations; for {\em independent} variable detection we use distances between local short-time dynamics.
We demonstrate each approach through an illustrative established PDE example. 
Such variable-free, emergent space identification algorithms connect naturally with
equation-free multiscale computation tools.

\end{abstract}

\section{INTRODUCTION}

For many systems of interest in physics or engineering, we are given a fine-scale description of the system evolution, e.g. at the  particle-based or agent-based level; yet the system exhibits large-scale, coarse-grained, spatiotemporal patterns which may well be captured by a set of unknown \emph{effective, coarse-grained} possibly \emph{emergent} PDEs. Such reduced, effective PDEs, when they exist and can be derived (whether mathematically, or in a data-driven fashion) can serve as cheap surrogate models, drastically facilitating computation-intensive tasks like prediction, optimization, uncertainty quantification and even control. 

A promising paradigm for the discovery of PDEs from data arises as part of the recent explosion of machine learning algorithms and their applications in the physical sciences. 
Since the early 90s, there has been interest in using neural networks in identifying nonlinear evolutionary PDEs from data (e.g. \cite{gonzalez1998identification}), and this interest has only grown in recent years to enlist other techniques, such as Gaussian Process Regression, Sparse Regression, Reservoir Computing and Recurrent/LSTM Networks \cite{rudy2017data,raissi2018numerical,pathak2018model,vlachas2018data}.  
Extensions of these techniques show great promise in the discovery of reduced, effective, coarse-grained and even emergent PDEs (a term we will explain below) {\em from fine-scale time series data} of distributed parameter systems \cite{kemeth2018emergent,bar2019learning,lee2020coarse,arbabi2020linking}. 

Despite its general promise, the machine-learning-based methodology typically comes with a nontrivial caveat, namely, the need to have \emph{a priori} knowledge about what the right space (independent) variables are, as well as the choice of the right observables (the right dependent variables) in terms of which the PDE can be written down, identified or learned. This problem is exacerbated when we are interested in discovering effective coarse-grained PDEs from data on individual agents or particles: these original microscopic degrees of freedom of the system are typically not good candidates for large-scale, collective, effective descriptions. In this work, we illustrate two frameworks, based on manifold learning, for discovering coarse-grained variables/observables directly from data.

In the first setup,  we assume that we have access to time-series data of (a large set of) interacting particles or agents, and the choice of independent variables (the right space-time) in the coarse-grained PDE is already known. For example, in the case of collective particle motion, a natural choice for such an independent variable would be the coordinates of the space in which the particles move, and the coarse-grained PDE would involve the spatial derivatives of some \emph{unknown, coarse} dependent variables. We assume that these unknown dependent variables capture the local collective (possibly averaged) statistical features  of the particles, and hence can be written in terms of the local particle distribution observations. We use manifold learning to extract these coarse nonlinear observables from mining local particle distributions, and illustrate how such coordinates can serve as dependent variables for a coarse-grained PDE. We demonstrate an application of this framework by coarse-graining a model of collective particle motion, known to macroscopically lead to the nonlinear Burgers PDE, in terms of a new, data-driven nonlinear dependent variable arising from processing the particle simulation data. 

In the second setup, we study problems where we have access to time-series data of a complex, many-degree-of-freedom system that we believe can be coarse-grained as a PDE, yet the right \emph{independent variable} (e.g.the right space coordinate) in which  to approximate the PDE is not known \emph{a priori}.
We discover useful unknown parameterizations (useful \emph{ emergent space coordinates}) by performing manifold learning on the local time-series data: We treat each time series as a single data point; then based on the assumption that, in the right embedding space, spatiotemporal proximity implies behavioral (time series) similarity, we identify nonlinear coordinates that exploit time-series similarity in order to usefully embed them. We then use this(ese) data-driven coordinate(s) as the independent variable(s) to formulate and accomplish the learning of the PDE. We illustrate this technique by relearning the complex Ginzburg-Landau PDE from time-series with deliberately scrambled spatial information. 

\section{Learning PDEs with unknown dependent variable(s)} \label{sec_coarse}

Consider a dynamical system composed of many interacting particles or agents where the evolution is prescribed at the microscopic, fine-scale level, i.e., we know the equations of motion for individual particles or agents.  These type of systems abound in engineering and the physical sciences, ranging from chemical reaction/transport occurring on lattices to the
coordinated collective motion of UAVs or living organisms. In many situations, such systems are known to give rise to large-scale, coarse-grained, coherent \emph{macroscopic} patterns and hence, it is reasonable to entertain the hypothesis that there exist macroscopic effective equations, e.g. in the form of PDEs, that can (approximately) predict the large-scale, coarse-grained system evolution. A critical step in discovering such macroscopic PDEs is finding the right collective observable(s) (e.g. the local density of particles, local particle pair and/or triplet  probabilities etc.) that can serve as the dependent variable(s) for the target PDE. Such unknown variables often have a statistical/averaged nature and therefore can be described in terms of the local agent distribution. In this section, we introduce a framework based on \emph{manifold learning using appropriate distances between local distribution observations}, leading to the discovery of suitable collective macroscopic variables from the data. For concreteness, we illustrate the framework through a particle-based model known to lead to a macro-scale PDE \cite{gear2003gap}.

Consider a system of moving particles, with the motion of each particle given by 
\begin{equation}\label{eq_Burgersmicro}
dZ= \frac{1}{2}\rho(Z,t) + \sqrt{2\nu}~d W,
\end{equation}
where $Z$ is the position of the particle in the periodic domain $[0,2\pi)$, $\rho(Z,t)$ is the local particle density at $Z$, $\nu$ is a (fixed) parameter of the motion and $W$ is the standard Wiener process. In this system, the particles are implicitly coupled through the local density term $\rho(Z,t)$. Our target is to find a PDE that usefully approximates the macro-scale, collective particle evolution. For this system, at an appropriate limit, such an equation exists: it is the nonlinear Burgers PDE,
\begin{equation}\label{eq_Burgersmacro}
 \p_t\rho = - \rho\p_z\rho + \nu \p_{zz} \rho,
\end{equation}
where $\rho$ is the particle density \emph{field}, the parameter $\nu$ (same parameter as in \eqref{eq_Burgersmicro}) is called viscosity, and $z \in[0,2\pi)$ is the spatial coordinate. As such, this equation describes the macroscopic evolution of the particle density field due to nonlinear advection (first term on the right-hand-side) and diffusion (second term).
We chose this example because of its nonlinearity --- an even simpler example would be, of course, the linear diffusion equation as collective description of the motion of random walkers with no interaction.

Consider the numerical simulation of the above system at the particle level and its connection with the macro-scale PDE.  The initial condition consists of many particles distributed over the domain, taken to correspond to some initial density profile, $\rho_0(z)$. We can define a resolution factor 
\begin{equation}\label{eq_resfactor}
 R = \frac{\text{total no. of particles}}{\int_{0}^{2\pi} \rho_0 dz},
\end{equation}
which describes the number of particles per unit mass. We assume $R$ is large enough so that at each time the local density can be meaningfully/robustly estimated at each spatial location in the domain: Let $w\ll 2\pi$  be the width of a small box centered around the position $z$. Then the particle density at $z$ can be approximated using the number of particles in that box, $n$, as
\begin{equation}\label{eq_density}
  \rho(z) \approx \frac{n}{w R}.
\end{equation}

\begin{figure}
\centering
{\includegraphics[scale=1]{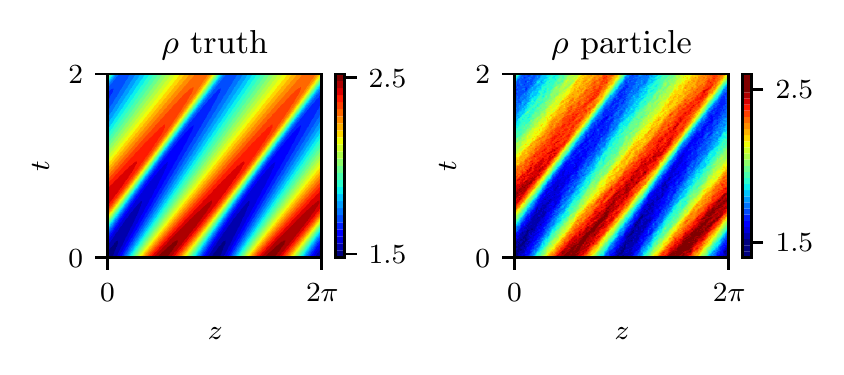}} 

\caption{A trajectory of the Burgers PDE with $\nu=0.05$ and initial condition $\rho_0(z)=1-\cos(2z)/2$ simulated via a high-order finite volume method (left) and via a corresponding microscopic system of interacting particles (right). For the micro simulations we have used $R=4\times 10^4$ and computed the density on a grid of size 128. The relative MSE of this particle simulation is $1.46\times 10^{-2}$.}
\label{fig_burgers}
\end{figure}

To evolve the system of particles, we march eq. \eqref{eq_Burgersmicro} forward in time using the Euler-Maruyama scheme \cite{higham2001algorithmic}. At each time step, we can compute the density profile using eq. \eqref{eq_density}. Conversely, given any density profile such as $\rho_0(z)$, we can generate a global distribution of particles consistent with that density profile using the particle cumulative distribution function, or through the following procedure: we partition the domain into many small boxes of width $w$ and compute the number of particles for that box from \eqref{eq_density}. We put that number of particles inside that box with positions drawn randomly from a uniform distribution.
Fig. \ref{fig_burgers} shows a trajectory of the PDE simulated with particle dynamics compared to the ``truth". The ``truth" in this case is a high-order finite-volume \cite{shu2009high} simulation of the Burgers PDE in eq. \eqref{eq_Burgersmacro}.

\subsection{Discovery of the coarse variable via manifold learning}
In the above example, we already know a good choice of macro-scale variable (i.e. the density field) for which a governing PDE exists. However, in many problems related to the macro-scale evolution of particle-based or agent-based systems, such a variable is not \emph{a priori} available. Here, we propose an approach for selecting a macro-scale variable from time-series of particle evolution data.  In our setup, it is anticipated that the macro-scale variable would encapsulate local statistics of the particles, and therefore we assume that the variable can be found in terms of the observed local distributions of particles. To capture this function, we use here the moments representation of the local particle distributions (other appropriate metrics, like an earth-mover distance, could also be used).
%

\begin{figure}
\centering
\subfloat[][]{\includegraphics[scale=1]{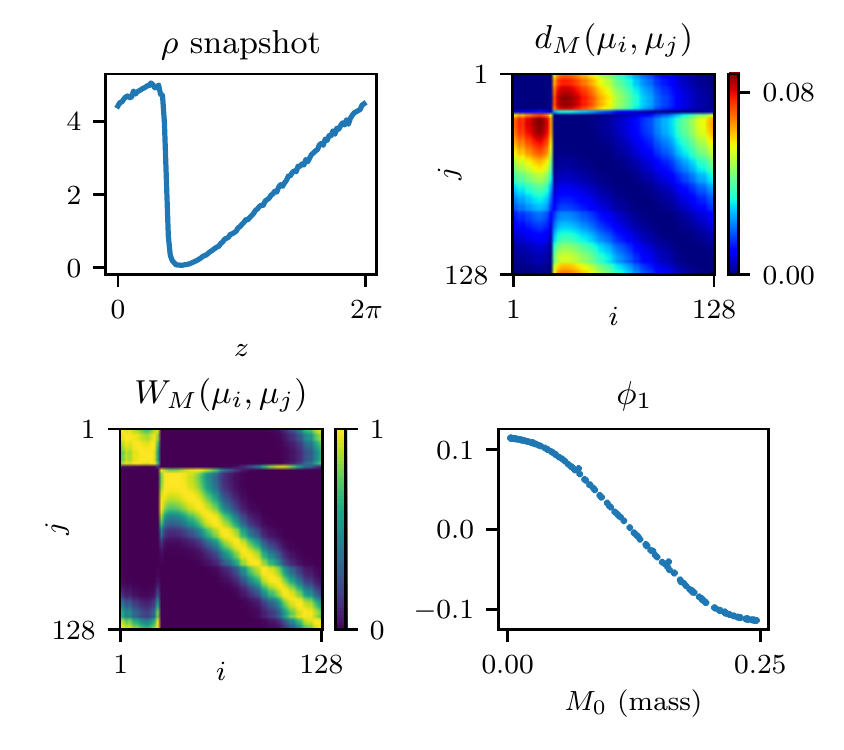}} 

\subfloat[][]{\includegraphics[scale=1]{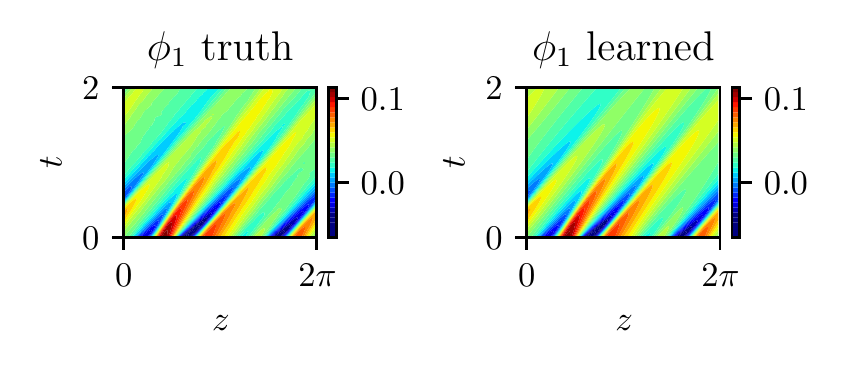}}

\caption{\textbf{Discovering the macro-scale variable and its governing PDE.} a) We take a snapshot of the particle simulation (top left), partition the domain into 128 boxes and compute the pairwise moment distance of particle position distributions across all box pairs (top right). We compute the diffusion kernel (bottom left) and the leading Diffusion-Maps coordinates. The only independent Diffusion-Maps coordinate, $\phi_1$, is one-to-one with the total mass within each box (and so, also, to the local density). b) Using a three-layer feedforward neural network we learn the PDE for $\phi_1$ and temporally integrate it along a test trajectory. The relative MSE of the trajectory is $2.71\times 10^{-2}$.  }
\label{fig_result1}
\end{figure}

Consider a snapshot of the particle system, consisting of the position of the particles at time $t$ (the corresponding density field is shown in Fig. \ref{fig_result1}(a, top left)).
We first partition the domain into a set of small equally sized boxes, and extract the local particle position distributions within each box. We can think of each  distribution as a {\em point in the space of distributions}, or a finite-dimensional approximation thereof and, therefore, the set of all observed distributions constitutes a cloud of data points in that space. We hypothesize that this observed point cloud lies close to a low-dimensional manifold. The coordinates of that manifold, which provide us with a parameterization of our data points, constitute candidates for the dependent variables of the coarse-grained PDE.

Capturing the nonlinear coordinates parameterizing this low-dimensional manifold can be accomplished through tools of manifold learning, here Diffusion Maps \cite{coifman2006diffusion}. The main ingredient required for the Diffusion Maps algorithm is a notion of distance (alternatively, a similarity measure) between pairs of data points. Given that here each data point is a particle distribution within a small box, we need to define a distance notion between such local distributions.

We use the moments representation of local observed particle distributions to define the distance in the space of distributions within our small boxes. 
Let $\mu$ denote the density of particles within a box. The $k$-th moment of $\mu$ is defined as 
\begin{align}
M_k(\mu)\approx\frac{1}{R}\sum_{p=1}^{n} x_p^k
\end{align}
where $x_p \in [0,1]$ is the (normalized) position of individual particles in the box. Given that the box is a compact set, the sequence of moments $\left\{M_k(\mu)\right\}_{k=0}^\infty$ uniquely distinguishes $\mu$ from other distributions \cite{schmudgen2017moment}. Here, we use a finite truncation of this sequence, i.e.,
\begin{align}
\tilde{M}(\mu) = \left[M_0(\mu),~M_1(\mu),\ldots,~M_K(\mu) \right]^\top.
\end{align}
to approximate the distance between particle distributions in the equally sized boxes.  
We define our \emph{moment distance} of the particle distributions within box $i$ and $j$, respectively denoted by $\mu_i$ and $\mu_j$, as the Euclidean distance between their truncated, raw moment vectors:
\begin{align}
d_M(\mu_i,\mu_j) = \left\| \tilde{M}(\mu_i) -\tilde{M}(\mu_j)\right\|.
\end{align}
This pairwise distance matrix for a snapshot of the particle simulation, using $K=6$ moments, is shown in Fig. \ref{fig_result1}(a). 
We observe that the qualitative structure of the distance matrix does not change when we vary the number of moments between 1 and 10.


To discover the appropriate data-driven descriptors we apply Diffusion Maps \cite{coifman2006diffusion} using the moments distance as our similarity measure in the diffusion kernel. We define the affinity kernel
\begin{align}
W_{ij}=\exp\left(-\frac{d_M(\mu_i,\mu_j)^2 }{\epsilon}\right)
\end{align}
where $\epsilon$ is the diffusion kernel width. Following standard practice, we choose $\epsilon$ to be the median of the distance values for our data points. We compute the normalized kernel matrix,
\begin{align}
\overline{{W}}= {D}^{-1}{W}{D}^{-1}
\end{align}
where ${D}$ is a diagonal matrix whose $i$-th entry is the sum of the $i$-th row in ${W}$. 
Next the kernel is made row-stochastic, i.e., 
\begin{align}
\hat{{W}}= \overline{{D}}^{-1}\overline{{W}}.
\end{align}
to obtain the (approximate) diffusion operator on the manifold,
\begin{align}
A={I}-\hat{{W}}.
\end{align}
 The diffusion map eigenfunctions are obtained in discretized form as eigenvectors of this matrix,
\begin{align}
{A}\phi_k=\lambda_k\phi_k, \quad k=0,1,\ldots,m.
\end{align}
 These eigenfunctions provide a set of (possibly dependent) coordinates for our data points
 (i.e. local spatial distributions).
 
 Fig. \ref{fig_result1}(a) shows the application of our framework to a snapshot of the particle simulation.  Analysis of the obtained eigenfunctions (not shown here) shows that there exists only a single independent Diffusion-Map coordinate for the observed data, denoted by $\phi_1$, and as expected,  this coordinate is in one-to-one correspondence with the total mass in each box (Fig. \ref{fig_result1}(a, bottom right)), which in turn, is proportional to the value of the local (inside the box) particle density. Note that we obtained the value of $\phi_1$ from each
 data snaphshot; we can find it for new snapshot observations using the Nystr{\"o}m extension; the diffusion map suite of tools also allows for function extensions on the manifold parameterized by $\phi_1$ through Geometric Harmonics \cite{coifman2006geometric}.

 \subsection{Discovering the macro-scale PDE using neural networks}
After discovering $\phi_1$ as the candidate for macro-scale dependent variable, we use a feedforward neural network (i.e. a multi-layer perceptron) to learn the PDE that governs its evolution.
We consider the following structural form for the discovery of the PDE that governs $\phi_1$, 
\begin{align} 
\p_t \phi_1=F(\phi_1,\p_z \phi_1,\p_{zz}\phi_1).
\end{align}
Once we have computed $\phi_1$ and its spatial derivatives for our particle simulation data, the learning problem is reduced to regressing the function $F$ from the data.
Note that the number of spatial derivatives required in the RHS is usually unknown beforehand and the data-driven technique introduced in \cite{lee2020coarse} can be used to search for that number.

The neural network  we use to regress $F$ has three hidden layers each consisting of 32 nodes. The first two layers are followed by rectifying linear unit activation functions. This type of architecture was successfully used in previous work \cite{arbabi2020linking} for coarse-graining PDEs with spatially heterogeneous parameters.
We implement the network in TensorFlow 2.0 \cite{tensorflow2015-whitepaper} and optimize it using the ADAM optimizer \cite{kingma2014adam}.

To generate training data for learning, we simulate eight two-second-long trajectories of the particle system with parameter values of $Z=4\times 10^4$ and $\nu=0.05$. The initial density of the particle simulations are given by
\begin{equation}\label{eq_IC}
\rho_0(x)=\sum_{k=1}^{20}A_k\sin(l_k x + \phi_k)
\end{equation}
where $A_k$, $l_k$ and $\phi_k$ are drawn randomly from uniform distributions on $[-.5,.5]$, $\{1,2,\ldots,6\}$ and $[0,2\pi)$, respectively. We transform the snapshots of particle simulations to the macro-scale variable $\phi_1$ and compute the associated values of $\p_t \phi_1$ using finite differences in time (more sophisticated algorithms finding continuous right-hand sides from discrete time data,
e.g. using Runge-Kutta inspired ResNets, can also be used). Using a sampling interval of $0.001$ for each trajectory, this results in 8000 pairs of $(\phi_1,\p_t \phi_1)$ global profiles to train the neural net.

We use the trained neural net to integrate a test trajectory with an unseen initial condition drawn from \eqref{eq_IC}. The results shown in Fig.~\ref{fig_result1}(b) corroborate that the learned PDE is sufficiently accurate and stable to yield accurate trajectories of the macro-scale evolution. 
Systematically translating from the coarse data-driven ($\phi_1$-based) description to a physically meaningful one is clearly a desirable feature of the overall process. Going from coarse $\phi_1$ to (also coarse) density can be performed, as mentioned above, through Geometric Harmonics. The more detailed, ``lifting" step from $\phi_1$ to particle position realizations has been discussed in equation-free computation \cite{kevrekidis2003equation}; it is worth noting that Generative Adversarial Networks (GANs) \cite{goodfellow2014generative} provide an alternative computational technology towards such a goal. 

\section{Learning PDEs with Unknown, Emergent  {\em Independent} Variables}
In many agent-based or particle-based systems, even the independent variables in which to first embed and then describe the coarse dynamics may be unknown. Consider for example social networks or systems of interacting neurons, where an arrangement of the agents or neurons in some physical space coordinate is not informative, and it is rather the connectivity that determines what an appropriate, ``emergent" embedding space might be.

In this section, we propose to extract the independent (space) variables of such a system in a data-driven way.
Based on the behavior of each agent, expressed for example through time series of their respective dynamic evolution variables,
we extract \textit{emergent} coordinates using Diffusion Maps~\cite{kemeth2018emergent}.
These emergent coordinates can then be used as independent variables in which we can learn effective evolutionary equations for multiagent systems.

\begin{figure}
\centering
\subfloat[][]{\includegraphics[scale=1]{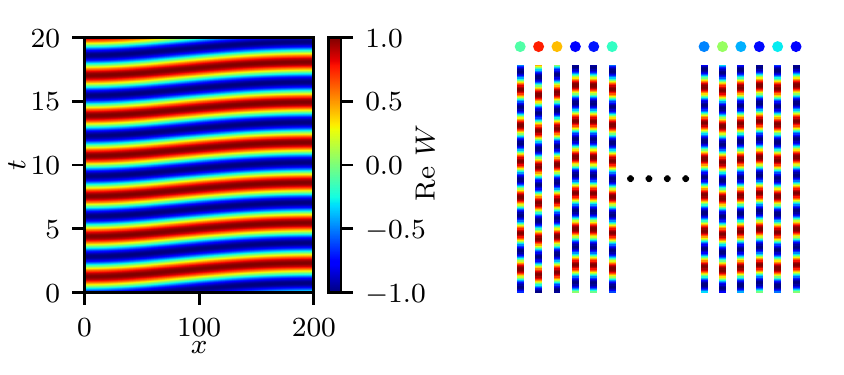}}

\subfloat[][]{\includegraphics[scale=1]{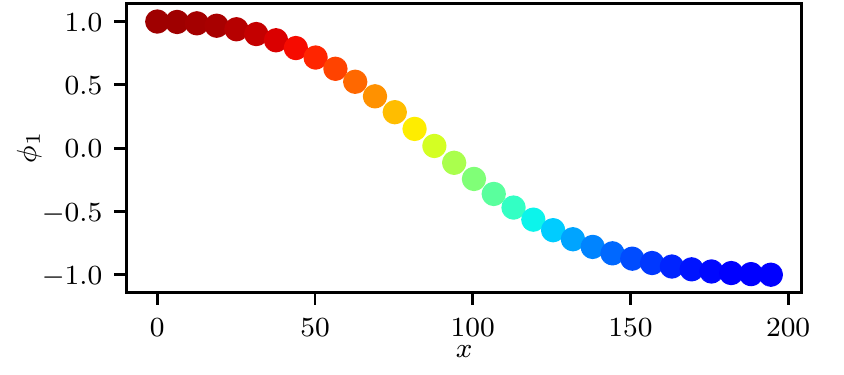}}

\subfloat[][]{\includegraphics[scale=1]{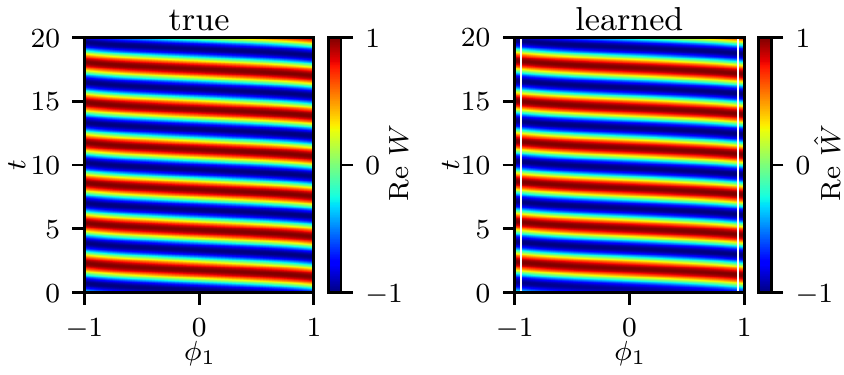}}

\caption{\textbf{Finding an effective PDE in emergent coordinates.} a) Simulation data of the complex Ginzburg-Landau equation with $N=128$ collocation points (left). 
The color indicates the real part of the complex field $W(x, t)$. 
Removing the spatial label of the $N=128$ collocation points leads to a collection 
of disorganized time series, each represented by a dot colored with the real parts of $W(t=20)$ on the right. b) Using the time series of (a) as input for the Diffusion-Maps algorithm leads to a one-dimensional Diffusion-Map coordinate $\phi_1$. Each entry corresponds to a single time series, and is colored with the respective Re $W(t=20)$ values. Plotting $\phi_1$ over the space coordinate $x$ reveals their one-to-one correspondence. For better visibility, only every fourth entry of $\phi_1$ is plotted.
(c) The data of (a), but now parameterized by the new, {\em emergent} space coordinate $\phi_1$ (left). Predictions of the learned PDE, of the same initial snapshot as shown on the left, are shown on the right.}
\label{fig_cgle}
\end{figure}

We illustrate this approach on an actual PDE example, the complex Ginzburg-Landau equation~\cite{garcia-morales12_compl_ginzb_landau_equat},
\begin{align}
\partial_t W(x, t) & = W(x, t) + \left(1+ic_1\right)\Delta W(x, t) \\ \nonumber
& - \left(1-ic_2\right)|W(x, t)|^2 W(x, t)
\label{eq:cgle}
\end{align}
describing the evolution of the complex field $W(x, t)$. 
Here we use the real-valued
parameters $c_1=1$, $c_2=2$, $L=200$ and zero-flux boundary conditions.
The numerical results of integrating the initial condition $W(x, 0) = (1+\cos(x\pi/L))/2$
are visualized in Fig.~\ref{fig_cgle}(a), where the color indicates the real part of $W(x,t)$
as a function of space and time. For integration,
a pseudo-spectral method with a fixed time step of $dt=0.01$ and exponential time stepping is used~\cite{Cox2002}.
The solution exhibits spatiotemporal periodic oscillation in time,
but with a different phase depending on the physical position $x$. Notice the flat regions near the boundaries,
originating from the zero-flux boundary conditions.

By discretizing the complex Ginzburg-Landau equation during numerical integration, the system can be viewed as an agent-based model. Here, each agent corresponds to one of the $N=128$ collocation points, and it is coupled to the neighboring agents through the discretization of the diffusion term.
Let us now {\em deliberately scramble the spatial location of the agents} (label them randomly, so that neighboring agents do not have consecutive labels). 
Not knowing the spatial location of these agents (i.e. the position of the collocation points), is there a way to
recover the fact that the system can be parameterized by a single spatial dimension, and can we discover the right embedding topology for the randomly labelled agents? 

In order to answer this question, we remove the spatial coordinate from each collocation point, resulting in a collection of $N=128$ randomly labelled ``shuffled" agents (cf. Fig.~\ref{fig_cgle}(a, right)).
In Fig.~\ref{fig_cgle}(a, right), the agents are indicated as points, colored with their real parts of $W(t=20)$.
Note that to each agent there ``belongs" a corresponding time series $W(t)$.
Regarding the time series of each agent as a data point, we apply Diffusion Maps as defined above. For the data considered here, one then finds a single independent coordinate $\phi_1$,
by which the time series can be parameterized. The resulting Diffusion-Map coordinate $\phi_1$ is depicted in Fig.~\ref{fig_cgle}(b) as a function of the actual physical space, where each entry of $\phi_1$ corresponds to one of the agents as shown in Fig.~\ref{fig_cgle}(a, right).
As can be observed from Fig.~\ref{fig_cgle}(b), the emergent coordinate $\phi_1$ is one-to-one with the physical space coordinate $x$. It is, however, flipped; that is, large $\phi_1$ values correspond to small $x$ and vice versa.
This can be explained through the ambiguity in the direction of the physical space $x$.
Furthermore, $\phi_1$ varies more strongly for agents based at intermediate values of $x$, indicating
that the agents vary more strongly in this interval, cf. Fig.~\ref{fig_cgle}(a, left).
Nevertheless, the right ``sequence of agents" - the right topology of the emergent embedding space - is preserved.
The data parameterized in this new data-driven coordinate $\phi_1$ is illustrated in Fig.~\ref{fig_cgle}(c, left). 
In this emergent coordinate, one is now able to learn an effective PDE description of the agents by learning an evolution equation of the form
\begin{equation}
  \partial_t W(\phi_1, t) = f\left(W, \frac{\partial W}{\partial \phi_1},
  \frac{\partial^2W}{\partial \phi_1^2},\frac{\partial^3W}{\partial \phi_1^3}\right),
\end{equation}
where $f$ is again represented by a neural network. Here, $f$ is composed of \(4\) fully connected hidden layers
with \(96\) neurons and bias terms each, followed by \(\mbox{tanh}\) nonlinearities.
The derivatives are obtained using finite differences with a stencil of width 9, 
%
%
using the highest order finite-difference approximation that fits into the stencil.
Since the data is not available at equidistant points in $\phi_1$, we resample them  on the interval $\left[-1, 1\right]$ at $N=128$ discrete points using a bivariate cubic spline.
The model $f$ is optimized by minimizing mean squared error between the temporal derivative 
$\partial_t W(\phi_1, t)$ and the predictions of the model, using the Adam optimizer~\cite{kingma2014adam} and a learning rate of $\lambda=10^{-3}$. For the remaining hyperparameters and weight initialization, the default settings in Pytorch are used~\cite{paszke2019pytorch}.
The model is trained for $60$ epochs on a training set of $8\times 10^5$ snapshots.
The snapshots thereby comprise $20$ transients approaching the limit cycle attractor,
thus also allowing $f$ to obtain information about the stability of the limit set.
See Ref.~\cite{kemeth2020_learning_pdes_in_emergent_coordinates} for details on the data sampling and
regularization.
In Fig.~\ref{fig_cgle}(c, right), predictions arising from integrating an initial snapshot from the test set using the learned model and forward
Euler with $dt=10^{-3}$ are shown. The recorded behavior over  boundary corridors is provided in lieu of boundary conditions during inference, as indicated by the white vertical thin lines. Notice the close correspondence of the true and learned dynamics, even when learned in a space coordinate in which no actual physical PDE is available.

\section{Discussion}
We presented two distinct ways of combining deep learning (in the form of neural networks) with manifold learning (in the form of Diffusion Map variations) in order to discover 
coarse-grained, effective partial differential operators (and the associated PDEs) from 
fine-scale simulation data. 
The data-driven discovery of {\em dependent} as well as {\em independent} variables was illustrated in separate examples, yet clearly the two modifications can be combined in a straightforward manner.

We have already mentioned that the approach naturally lends itself to links with modern multiscale numerical algorithms.
In a recent publication \cite{arbabi2020linking} we demonstrated how equation-free multiscale numerics (and, in particular, gap-tooth and patch dynamics schemes) can be used to substantially reduce the computational effort required to collect the fine-scale training data we need for our coarse-grained PDE identification.
This is accomplished by judiciously performing the fine-scale simulation during the data-collection phase in a relatively small fraction of the space-time domain \cite{kevrekidis2003equation}.
Translating between data-driven and physically meaningful representations of the simulation
(the ``lifting" and ``restriction" operations of equation-free computations) were addressed here through manifold learning tools (Nystr{\"o}m extension and Geometric Harmonics), although alternative techniques (e.g. Gaussian Processes) can also be used.
Our emergent space example here was selected to be an instructional one: the known physical space was scrambled, and, then, in a sense as in the case of a puzzle, recovered. We are currently experimenting with this approach in cases where a meaningful spatial description is not known {\em a priori} to exist, including large {\em in silico} networks of biological neurons, as well as social and  power networks with complex connectivity (\cite{kemeth2020_learning_pdes_in_emergent_coordinates}). 
\addtolength{\textheight}{-0cm}   

%


\bibliography{root.bbl}
\bibliographystyle{abbrv}

\end{document}